\documentclass[10pt,twocolumn,letterpaper]{article}

\usepackage{cvpr}              % To produce the CAMERA-READY version
\definecolor{cvprblue}{rgb}{0.21,0.49,0.74}
\usepackage[pagebackref,breaklinks,colorlinks,allcolors=cvprblue]{hyperref}
\usepackage{algorithm}
\usepackage{algpseudocode}
\usepackage{multirow}
\usepackage{graphicx}
\usepackage{booktabs}
\usepackage{lscape}
\usepackage{colortbl}
\usepackage{pifont}
\usepackage{fixltx2e}
\usepackage[normalem]{ulem}
\usepackage{makecell}

\usepackage{xcolor}
\usepackage{soul} % for \sethlcolor and \hl

\definecolor{neonred}{RGB}{255, 153, 153}     % 옅은 형광빨강 (soft neon red)
\definecolor{neongreen}{RGB}{153, 255, 153}   % 옅은 형광초록 (soft neon green)

\definecolor{brickred}{rgb}{0.8, 0.25, 0.33}
\definecolor{brickgreen}{rgb}{0.25, 0.8, 0.33}
\newcommand{\cm}{\textcolor{brickgreen}{\ding{51}}}%
\definecolor{HeaderBlue}{RGB}{233,242,255}
\definecolor{RowGray}{RGB}{246,246,246}
\definecolor{SOTARed}{RGB}{215,35,45}
\definecolor{PrevBlue}{RGB}{42,96,255}
\definecolor{ConfCyan}{RGB}{0,153,204}

\newcommand{\sota}[1]{\textcolor{SOTARed}{\bfseries #1}}
\newcommand{\prevsota}[1]{\textcolor{PrevBlue}{#1}}
\newcommand{\conf}[2]{\textcolor{ConfCyan}{\textsc{#1}\,#2}}
\newcommand{\confsub}[2]{\textsubscript{\conf{#1}{#2}}}

\def\paperID{76
91} % *** Enter the Paper ID here
\def\confName{CVPR}
\def\confYear{2026}

\title{
\raisebox{-0.7em}{\includegraphics[width=1.2cm]{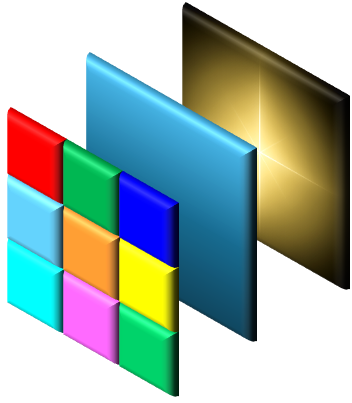}}\hspace{0.4em}
UCMNet: Uncertainty-Aware Context Memory Network for Under-Display Camera Image Restoration
}

\author{
Daehyun Kim$^1$, Youngmin Kim$^{1,2}$, Yoon Ju Oh$^1$, Tae Hyun Kim$^{1}$\thanks{Corresponding author}\\
Hanyang University$^1$, Agency for Defense Development (ADD)$^2$\\
{\tt\small \{daehyun, youngmin4507, oyj0813, taehyunkim\}@hanyang.ac.kr}
}

\begin{document}
\maketitle

\begin{abstract} 
Under-display cameras (UDCs) allow for full-screen designs by positioning the imaging sensor underneath the display. Nonetheless, light diffraction and scattering through the various display layers result in spatially varying and complex degradations, which significantly reduce high-frequency details. 
Current PSF-based physical modeling techniques and frequency-separation networks are effective at reconstructing low-frequency structures and maintaining overall color consistency. However, they still face challenges in recovering fine details when dealing with complex, spatially varying degradation.
To solve this problem, we propose a lightweight \textbf{U}ncertainty-aware \textbf{C}ontext-\textbf{M}emory \textbf{Network} (\textbf{UCMNet}), for UDC image restoration. 
Unlike previous methods that apply uniform restoration, UCMNet performs uncertainty-aware adaptive processing to restore high-frequency details in regions with varying degradations.
The estimated uncertainty maps, learned through an uncertainty-driven loss, quantify spatial uncertainty induced by diffraction and scattering, and guide the Memory Bank to retrieve region-adaptive context from the Context Bank. 
This process enables effective modeling of the non-uniform degradation characteristics inherent to UDC imaging. 
Leveraging this uncertainty as a prior, UCMNet achieves state-of-the-art performance on numerous benchmarks with 30\% fewer parameters than previous models. 
Project page: \href{https://kdhrick2222.github.io/projects/UCMNet/}{https://kdhrick2222.github.io/projects/UCMNet}. 
\end{abstract}

\begin{figure}[h]
    \centering
    \includegraphics[width=\columnwidth]{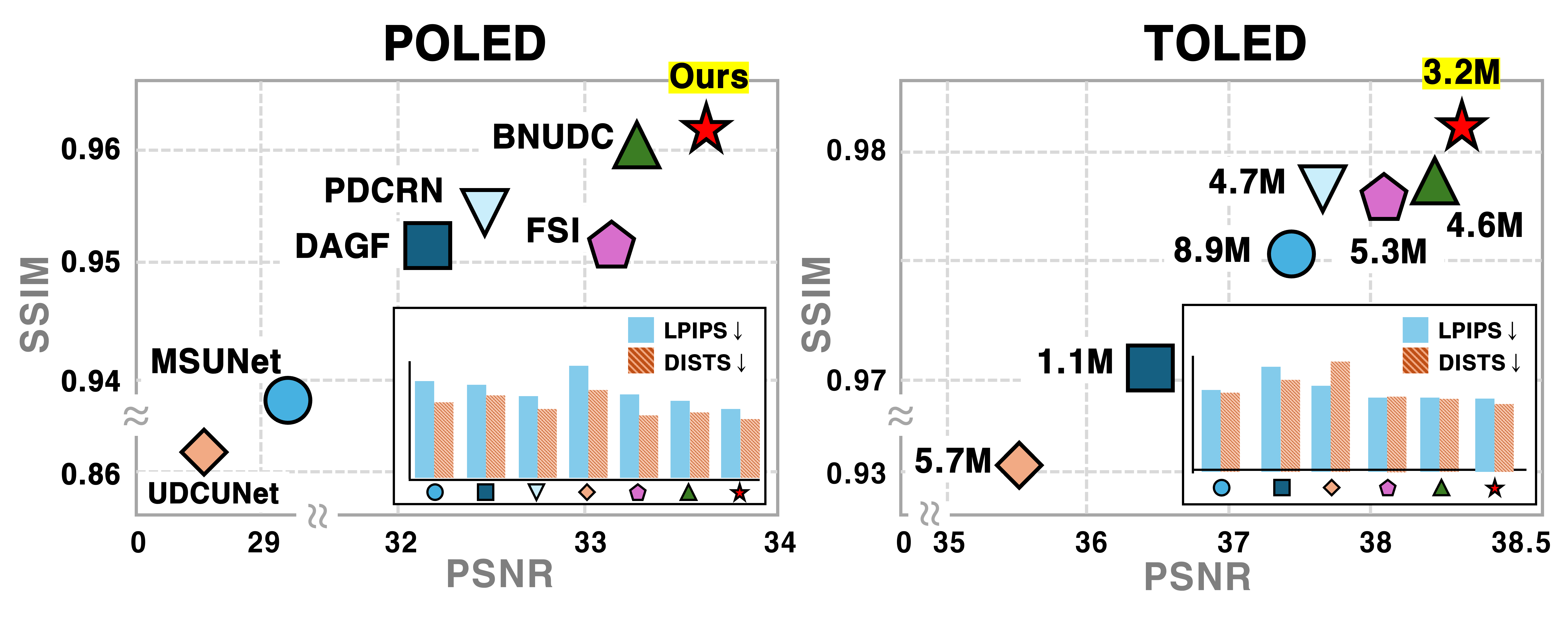}
    \caption{
    PSNR/SSIM comparisons on the POLED and TOLED datasets.
    Each marker denotes a restoration model positioned by its PSNR (x-axis) and SSIM (y-axis).
    \textbf{UCMNet} lies in the upper-right region, delivering the best performance and computational efficiency among all competing methods.
    }
    \label{fig:concept}
\end{figure}

%% 그림 고민중

\begin{figure*}[]
    \centering
    \includegraphics[width=1.0\linewidth]{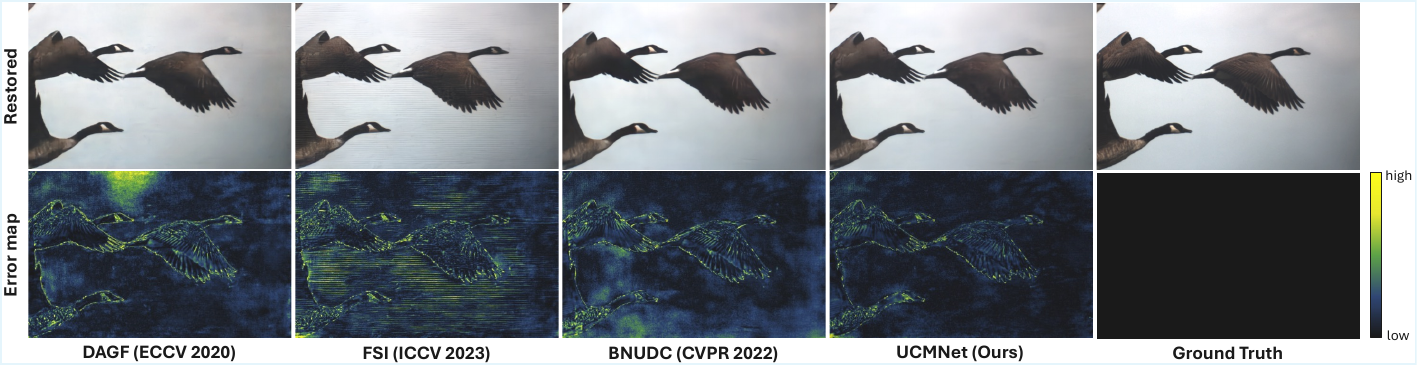}
    \caption{\textbf{Visual comparison of restored results (top row) and error maps (bottom row)} among existing UDC restoration models and the proposed UCMNet. UCMNet shows fewer artifacts and more accurate texture reconstruction (blue: small errors, yellow: large errors).}
    \label{fig:problem_example}
\end{figure*}

\section{Introduction} 
Full-screen mobile devices enhance user immersion by maximizing screen-to-body ratio and removing visible camera holes or notches, marking a significant step toward seamless visual experiences. 
However, positioning the front-facing camera beneath the OLED display 
introduces severe degradations through diffraction, scattering, and 
internal reflection, resulting in reduced transmittance, blur, noise, 
and flare.
These 
degradations exhibit spatial heterogeneity in two aspects: they vary 
across different panel types~\cite{kim202118,lim202074,qin2021p,zhang202114,zhang2020image}
, and more critically, 
within individual images where the lens center and boundary regions 
show vastly different diffraction characteristics, leading to 
location-dependent degradation patterns. This makes UDC image 
restoration a complex challenge that simultaneously involves denoising~\cite{li2024synthetic,Cheng_2024_CVPR,Ma_2025_CVPR}, 
deblurring~\cite{Wu_2024_CVPR,Youk_2024_CVPR,sun2024motion,yang2025gyro}, low-light enhancement~\cite{he2025retidiff,yan2025hvi}, and spatially-adaptive restoration.

Recent studies on UDC image restoration have explored various approaches to solve this problem, including physics-guided modeling~\cite{zhou2021image,DISCNet,kwon2021controllable,yang2021designing}, joint learning frameworks~\cite{kim2021under,zhou2022modular,gao2021image,song2023under}, and frequency-domain methods~\cite{panikkasseril2020transform,koh2022bnudc,liu2023fsi}. 
Although these techniques can partially recover high-frequency details of the captured scene, they still struggle with spatially varying artifacts caused by the display itself, resulting in unresolved residual distortions. 

In this work, we propose an effective network architecture that surpasses previous methods, as shown in Fig.~\ref{fig:concept}, by addressing spatially varying degradations caused by OLED-induced light disturbance through a novel training strategy tailored for UDC image restoration from an uncertainty-driven perspective. 
As demonstrated in the error map comparisons in Fig.~\ref{fig:problem_example}, conventional methods exhibit limitations not only in reconstructing object structures but also in handling UDC-specific artifacts, such as grid patterns, whereas the proposed method effectively overcomes these limitations. 
First, inspired by~\cite{ning2021uncertainty,chen2023sparse,chen2023uncertainty}, we investigate the potential of an approach guided by uncertainty. In this context, uncertainty measures the confidence level of predictions on a pixel-by-pixel basis and indicates the ambiguity found in visually deteriorated areas.
In contrast to deterministic methods that consider all pixels the same, uncertainty-aware modeling enables the network to distinguish between areas of high and low error. This method facilitates an improved focus on locally differently degraded regions, which commonly occur in UDC images.
Therefore, this uncertainty guided perspective is integrated into our network and learning framework to enhance restoration under complex, spatially varying degradations. 
Next, to improve the recovery of high-frequency details in UDC images, we present a high-frequency uncertainty-driven loss (HF-UDL), which facilitates the estimation of more precise uncertainty in high-frequency areas. This method enables the model to more effectively capture intricate structural details compared to the traditional uncertainty-driven loss (UDL).
Lastly, we introduce a new Uncertainty-Prior Transformer (UPT) to leverage the precisely estimated uncertainty maps. 
Specifically, within the UPT, we employ two memory banks (Memory Bank and Context Bank): the Memory Bank learns and stores uncertainty patterns, while the Context Bank learns and stores high-frequency information suitable for those specific patterns. 
Essentially, the Memory Bank acts as a key to the Context Bank. At each pixel location, we find the address in the Memory Bank that matches the uncertainty pattern of an input UDC image. This address then points to the Context Bank, allowing us to retrieve the learned high-frequency information. 
We refer to this model as the \textbf{U}ncertainty-aware \textbf{C}ontext-\textbf{M}emory \textbf{Network} (\textbf{UCMNet}), and our main contributions are outlined as follows:
\begin{itemize}
    \item We introduce a novel uncertainty-driven framework for UDC image restoration. 
    \item We design a high-frequency uncertainty-driven loss for effective uncertainty learning and refines fine details. 

    \item We propose an Uncertainty-Prior Transformer that leverages uncertainty maps to adaptively restore spatially varying degradations. 

    \item As shown in Fig.~\ref{fig:concept} and Fig.~\ref{fig:problem_example}, UCMNet sets a new standard in UDC image restoration, achieving state-of-the-art results on POLED~\cite{zhou2021image}, TOLED~\cite{zhou2021image}, and SYNTH~\cite{DISCNet} benchmarks.
\end{itemize}

\begin{figure*}[]
    \centering
    \includegraphics[width=1.0\textwidth]{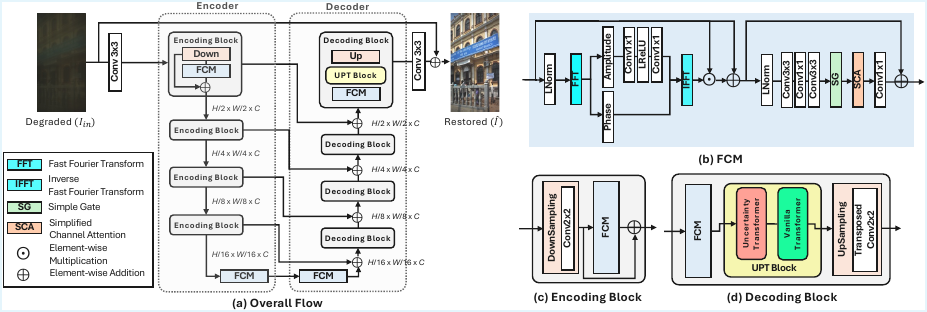}
    \caption{Architecture of the proposed method for UDC image restoration. Our model follows a U-shaped encoder–decoder architecture. The core module of the encoding block is the Frequency Convolution Module (FCM), while the decoding block additionally incorporates the Uncertainty Prior Transformer (UPT) block for uncertainty-guided feature refinement.}
    \label{fig:network_overall}
\end{figure*}

\section{Related Works}

\subsection{UDC Image Restoration}
Under-display camera (UDC) image restoration has recently emerged as a challenging real-world image restoration problem, characterized by complex, compound degradations. This requires concurrently addressing multiple tasks, including denoising, deblurring, and super-resolution.
Numerous deep neural networks are proposed to tackle this issue. 
Zhou \textit{et al.}~\cite{zhou2021image} are the first to introduce a physical image formation model for the UDC degraded image. 

Feng~\textit{et al.}~\cite{DISCNet} propose a dynamic skip connection network that leverages PSF priors to remove diffraction artifacts and introduce a synthetic data generation strategy tailored for UDC restoration. 
Similarly, a pixel-wise controllable architecture is introduced in ~\cite{kwon2021controllable} to address spatially variant diffraction blur and noise in an controllable manner, and in ~\cite{sundar2020deep}, an approach based on guided filter is proposed for local detail recovery. 
Additionally, Yang~\textit{et al.}~\cite{yang2021designing} investigate invertible blur kernels via Fourier optics. 

Another line of research tackles the UDC problem through joint learning frameworks, which simultaneously address multiple degradation factors. 
Kim~\textit{et al.}~\cite{kim2021under} jointly perform demosaicking and deblurring to reconstruct Quad Bayer RAW images for restoration, while Zhou~\textit{et al.}~\cite{zhou2022modular} integrate a GAN-based degradation simulator to train a network.
Gao~\textit{et al.}~\cite{gao2021image} propose a two-stage pipeline combining synthetic dataset construction and coarse-to-fine restoration, while Song~\textit{et al.}~\cite{song2023under} enhance the UDC imaging model by incorporating scattering and contrast distortion to generate more realistic training data. 

Recent studies~\cite{xu2020learning,mao2023intriguing,zhuang2022dpfnet} have explored the frequency characteristics of diffraction artifacts, aiming to complement frequency-aware representations. 
\cite{panikkasseril2020transform} address UDC degradation in the wavelet space through multi-scale frequency decomposition, while BNUDC~\cite{koh2022bnudc} adopts a two-branch network to separately restore low- and high-frequency components. Building on these ideas, FSI~\cite{liu2023fsi} introduces a frequency and spatial interactive network that jointly learns in both domains, highlighting that frequency-domain priors can enhance robustness to complex diffraction effects. 
However, previous methods overlook the spatially varying uncertainty in diffraction-induced degradations, whereas our approach integrates an uncertainty-driven mechanism to effectively restore non-uniform degradations present in real-world UDC images.  

\subsection{Uncertainty in Image Restoration}
Uncertainty modeling has emerged as a powerful tool for enhancing the reliability of deep neural networks in pixel-level prediction tasks. 
Kendall~\textit{et al.,}~\cite{kendall2017uncertainties} first formulate a Bayesian deep learning framework that jointly captures both aleatoric and epistemic uncertainties, demonstrating its effectiveness in semantic segmentation and depth regression. 
Then, Chang~\textit{et al.,}~\cite{chang2020data} further investigate data uncertainty by estimating the predictive mean and variance in face recognition. 

This probabilistic formulation from ~\cite{kendall2017uncertainties, chang2020data} is later extended to single-image super-resolution tasks in~\cite{ning2021uncertainty}, giving rise to an uncertainty-driven loss (UDL) defined as follows:

\vspace{-10px}
\begin{equation}
    \mathcal{L}_{\text{UDL}}= \exp(-s)\,\|\hat{I} - I_{gt}\|_1+ 2s,
    \label{eq:Loss3}
\end{equation}
where $s$ denotes the uncertainty estimate, and $\hat{I}$ and $I_{gt}$ represent the restored and ground-truth image. 
The concept of exploiting uncertainty as prior knowledge has been extended beyond its original contexts to diverse low-level vision tasks. 
Recent studies have incorporated uncertainty into low-light image enhancement~\cite{hou2023global} and all-in-one image restoration models~\cite{wu2025debiased}, where it serves as a regularizing prior for noise control within diffusion-based frameworks. 
In parallel, deraining and desnowing~\cite{chen2023sparse,chen2023uncertainty} networks have embedded uncertainty directly into their loss formulations, introducing sparse sampling and rank-based matrix constraints that further expand the methodological landscape of image restoration. 
 
Inspired by recent advances, we design a novel High-Frequency Uncertainty-Driven Loss (HF-UDL) to capture and utilize uncertainty arising from light diffraction inherent in UDC imaging.
This enables a more systematic regulation of the restoration process, even under diffraction-induced degradations. 

\begin{figure*}[h]
    \centering
    \includegraphics[width=1.0\linewidth]{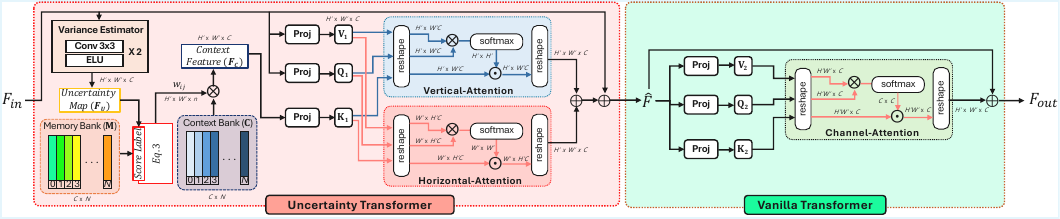}
    \caption{
    \textbf{Architecture of the Uncertainty-Prior Transformer (UPT) block.}
    The uncertainty transformer refines the input feature $F_{in}$ by predicting an uncertainty map and retrieving context features from the Memory and Context Banks, producing the uncertainty-enhanced representation $\hat{F}$ through vertical–horizontal cross-attention. 
    A subsequent vanilla transformer applies channel-wise self-attention to yield the final output $F_{out}$, improving restoration in uncertain and high-frequency regions.
    }
    \label{fig:transformer}
\end{figure*}

\section{Proposed Method}
In Fig.~\ref{fig:network_overall} (a), we illustrate the overall architecture of the proposed method.
A degraded UDC image $I_{in} \in \mathbb{R}^{H \times W \times 3}$ where $H$ and $W$ denote the height and width of the input image, is fed into a U-shaped residual network.
The network consists of an encoder with multiple encoding blocks that effectively extracts feature embeddings with downsampling, a decoder with decoding blocks that refines and enhances the embedded features while progressively upsampling them, and a Frequency Convolution Module (FCM) placed within encoding and decoding blocks to enhance features in the Fourier domain.  

\subsection{Encoder and Decoder} 
In the encoding process, the UDC input $I_{in}$ is transformed into a high-dimensional feature that contains abundant information. 
Initially, the encoder utilizes a convolutional layer on the UDC input to derive low-level feature embeddings denoted as $F_0 \in \mathbb{R}^{H \times W \times C}$, where $C$ represents the number of feature channels. This process is followed by multiple encoding blocks, and generates downsampled feature representations that preserve essential structural information for subsequent restoration in the decoder. 

In the decoding process, encoded features are progressively transformed back into a restored UDC image with fine-grained details and high-frequency components. 
Initially the decoder employs multiple decoding blocks that iteratively recover structural and textural information from the encoded features. 

\subsubsection{Frequency Convolution Module (FCM)}
In Fig.~\ref{fig:network_overall} (b), we introduce Frequency Convolution Module (FCM) designed to actively capture and address degradations occurring due to the under-display condition by explicitly utilizing information from the Fourier domain.
Inspired by NAFBlock~\cite{chen2022nafnet,feijoo2025darkir},
our FCM consists of a frequency domain feature enhancement part and a spatial attention part, with additional implementation details provided in the supplementary material Sec. A.

\subsubsection{Encoding Block} 
As illustrated in Fig.~\ref{fig:network_overall} (c), 
the encoding block downsamples the input using a 2$\times$2 convolution with stride 2 to reduce spatial resolution and expand the receptive field, while the FCM supports the encoding by integrating enhanced frequency-domain features. 

\subsubsection{Decoding Block}
As shown in Fig.~\ref{fig:network_overall} (d), the decoding block consists of three components: FCM, an Uncertainty-Prior Transformer (UPT), and an upsampling. 
Initially, input feature of the decoding block is enhanced by the FCM, which strengthens high-frequency representations in the Fourier domain.
Subsequently, UPT adaptively reconstructs spatially varying degradations resulting from display-induced diffraction and scattering by utilizing uncertainty-guided context. 
Lastly, a transposed convolution layer upsamples the refined features, restoring their spatial resolution.

\subsection{Uncertainty-Prior Transformer Block}
The Uncertainty-Prior Transformer (UPT) block, located within the decoding block, is the core module of our UCMNet and is introduced to leverage the uncertainty map in a spatially adaptive manner.
In the UPT block, learned high-frequency context features are retrieved from memory banks to enhance the input feature in an uncertainty-prior manner. 
These context features are then fused with the input via directional cross-attentions, followed by a vanilla transformer that enforces global coherence through channel-wise self-attention. 
Significantly, the UPT block facilitates locally adaptive refinement, playing a vital role in UDC image restoration.

\subsubsection{Uncertainty-Prior}
\label{subsubsec:uncertainty_prior}

As shown in the Uncertainty Transformer in Fig.~\ref{fig:transformer}, for the input feature $F_{in}\in\mathbb{R}^{H' \times W' \times C}$, a variance estimator initially predicts an uncertainty map $F_U\in\mathbb{R}^{H' \times W' \times C}$, which emphasizes features with high uncertainty. 
This uncertainty prior is then used to retrieve restoration-relevant information from two learnable memory banks:
a Memory Bank $\mathbf{M}$ and a Context Bank $\mathbf{C}$, defined as:
\begin{equation}
    \mathbf{M} = [m_1, m_2, \ldots, m_N], \quad \mathbf{C} = [c_1, c_2, \ldots, c_N],
\end{equation}
where $\mathbf{M} \in \mathbb{R}^{N 	\times C}$ and $\mathbf{C} \in \mathbb{R}^{N 	\times C}$ consist of $N$ learnable token pairs. Each memory token $m_i \in \mathbb{R}^{1 \times C}$ learns and stores uncertainty patterns. The corresponding context token $c_i \in \mathbb{R}^{1 	\times C}$ learns and stores compensating high-frequency information.

From the given uncertainty map, to select appropriate context tokens for each pixel location, we first compute the cosine similarity between each memory token $m_i$ and the uncertainty feature vectors $f^u_j \in \mathbb{R}^{1\times C}, j=1, \ldots, H'W'$ from the uncertainty map $F_U$ as:
\vspace{-5px}
\begin{equation}
    s_{ij} = \frac{m_i {f^u_j}^\top}{\|m_i\|\cdot\|f^u_j\|}.
    \label{eq:cosinesim}
\end{equation}
Then we apply the softmax function to the cosine similarity scores and obtain $w_{ij}$ which denotes the weight of the context token $c_i$.
Using the association weights $w_{ij}$, we retrieve context feature vector $f^c_j$, computed by aggregating the corresponding context tokens as $f^c_j = \sum^N_i w_{ij} c_i$, where $f^c_j$ denotes the $j$-th feature vector of $F_C$.
By utilizing paired Memory-Context Banks, UPT produces $F_C$, which signifies the high-frequency context feature retrieved for local degradation patterns, thus enhancing restoration in uncertain regions.

\subsubsection{Directional Cross-Attention}

In Uncertainty Transformer, to fuse the retrieved context with the input feature, we apply a cross-attention mechanism. 
The input feature $F_{in}$ and context feature $F_C$ are linearly projected to obtain the query, key, and value representations 
$\mathbf{Q}_1, \mathbf{K}_1, \mathbf{V}_1 \in \mathbb{R}^{H' \times W' \times C}$ using projection matrices 
$W_{\mathbf{Q}_1}, W_{\mathbf{K}_1},$ and $W_{\mathbf{V}_1}$ as:
\begin{equation}
    \scalebox{0.95}{$
        \mathbf{Q}_1 = F_{in} W_{\mathbf{Q}_1}^{\top}, \quad 
        \mathbf{K}_1 = F_C W_{\mathbf{K}_1}^{\top}, \quad 
        \mathbf{V}_1 = F_{in} W_{\mathbf{V}_1}^{\top}.
    $}
\end{equation}
We use $F_C$ as the key to guide restoration, while the query and value are taken from $F_{in}$ to preserve its spatial content.

To enable directional modeling of spatial dependencies, vertical and horizontal attentions are applied in parallel. 
For vertical attention, $\mathbf{Q}_1, \mathbf{K}_1, \mathbf{V}_1 \in \mathbb{R}^{H' \times W' \times C}$ are reshaped to 
$\mathbf{Q}_v, \mathbf{K}_v, \mathbf{V}_v \in \mathbb{R}^{H' \times (W'C)}$, 
and for horizontal attention, they are reshaped to 
$\mathbf{Q}_h, \mathbf{K}_h, \mathbf{V}_h \in \mathbb{R}^{W' \times (H'C)}$ and yield:
\begin{equation}
    \scalebox{0.9}{$
    F_v =\text{softmax}\!\left(\frac{\mathbf{Q}_v\mathbf{K}_v^{\top}}{\sqrt{\alpha}}\right)\mathbf{V}_v, 
    \quad
    F_h =\text{softmax}\!\left(\frac{\mathbf{Q}_h\mathbf{K}_h^{\top}}{\sqrt{\alpha}}\right)\mathbf{V}_h,
    $}
\label{eq:hvattention}
\end{equation}
where $F_v$ and $F_h$ denote the vertically and horizontally attended features, respectively, and $\alpha$ is a learnable scaling parameter. 
This parallel design enables the model to jointly exploit vertical and horizontal dependencies, leading to balanced spatial feature refinement. 
After reshaping $F_v$ and $F_h$ to $\mathbb{R}^{H' \times W' \times C}$, we obtain the output of Uncertainty Transformer as follows:
\begin{equation}
    \hat{F} = 0.5\times(F_v + F_h) + F_{in}. 
    \label{eq:Fhat}
\end{equation}

\subsubsection{Vanilla Transformer}

To complement local refinement and enforce global consistency, the uncertainty-enhanced feature $\hat{F}$ is further processed by a Vanilla Transformer~\cite{vaswani2017attention} that models long-range channel dependencies. 
As shown in Fig.~\ref{fig:transformer} (right), $\hat{F}$ is linearly projected to obtain 
$\mathbf{Q}_2, \mathbf{K}_2, \mathbf{V}_2 \in \mathbb{R}^{H' \times W' \times C}$ using projection matrices 
$W_{\mathbf{Q}_2}, W_{\mathbf{K}_2},$ and $W_{\mathbf{V}_2}$, and yields:
\begin{equation}
    \mathbf{Q}_2 = \hat{F} W_{\mathbf{Q}_2}^{\top}, \quad
    \mathbf{K}_2 = \hat{F} W_{\mathbf{K}_2}^{\top}, \quad
    \mathbf{V}_2 = \hat{F} W_{\mathbf{V}_2}^{\top}.
\end{equation}
Then, self-attention is applied to aggregate global context:
\begin{equation}
\scalebox{0.9}{$
\begin{aligned}
    F_{out} &= \text{Attn}(\mathbf{Q}_2, \mathbf{K}_2, \mathbf{V}_2) + \hat{F}, \\
    \text{Attn}(\mathbf{Q}_2, \mathbf{K}_2, \mathbf{V}_2) &=
     \text{reshape}\!\left(
        \text{softmax}\!\left(
            \frac{\mathbf{Q}_2\mathbf{K}_2^{\top}}{\sqrt{\beta}}
        \right)
        \mathbf{V}_2
     \right),
\end{aligned}
$}
\label{eq:vanilla_attention}
\end{equation}
where $\beta$ is a learnable positive scaling parameter, and $\text{reshape}$ denotes an operation that transforms the feature from $\mathbb{R}^{H’W’\times C}$ to $\mathbb{R}^{H’\times W’\times C}$. 
By globally processing the uncertainty-prior enhanced feature $\hat{F}$ through this self-attention mechanism, the Vanilla Transformer consolidates long-range dependencies and refines the feature representation, yielding the final output $F_{out}$ with improved structure and restoration quality.

\begin{figure}[hbt]
    \centering
    \includegraphics[width=1.0\linewidth]{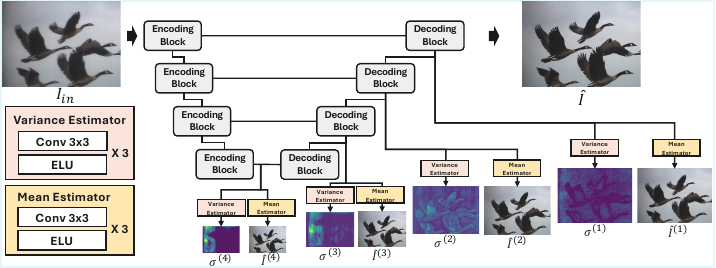}
    \caption{
    Uncertainty maps are derived from the uncertainty-driven loss, where each decoding block includes parallel mean and variance estimators that jointly predict the restored image and its corresponding uncertainty map. 
    }
    \label{fig:UDLoss}
\end{figure}

    \renewcommand{\arraystretch}{1.25}
    \begin{table*}[h]
    \small
    \centering
    \caption{
    \textbf{Quantitative results on the POLED and TOLED test datasets}. 
    Average PSNR, SSIM, LPIPS, and DISTS are reported, with the 
    \sota{best} and \prevsota{second-best} scores colored 
    ($\uparrow$ higher is better; $\downarrow$ lower is better).
    }
        \resizebox{\textwidth}{!}{%
    \begin{tabular}{l|cc|cccc|cccc}
    \toprule[0.5pt]
    \rowcolor{HeaderBlue}
    \multirow{2}{*}{Method} & \multirow{2}{*}{\makecell[c]{\#Params\\(M)}} & \multirow{2}{*}{\makecell[c]{FLOPs\\(G)}}  & \multicolumn{4}{c|}{POLED-Test~\cite{zhou2021image}} & \multicolumn{4}{c}{TOLED-Test~\cite{zhou2021image}} \\
    \rowcolor{HeaderBlue}
    \multirow{-2}{*}{Method} & \multirow{-2}{*}{\makecell[c]{\#Params\\(M)}} & \multirow{-2}{*}{\makecell[c]{FLOPs\\(G)}}   & PSNR$\uparrow$ & SSIM$\uparrow$ & LPIPS$\downarrow$ & DISTS$\downarrow$ & PSNR$\uparrow$ & SSIM$\uparrow$ & LPIPS$\downarrow$ & DISTS$\downarrow$ \\
    \midrule[0.5pt]
    MSUNet\confsub{CVPR}{2021}~\cite{zhou2021image} & 8.9 & 124.36 & 29.17 & 0.9393 & 0.2239 & 0.1746 & 37.40 & 0.9756 & 0.1093 & 0.1052 \\
    DAGF\confsub{ECCV}{2020}~\cite{sundar2020deep} & 1.1 & 36.77 & 32.29 & 0.9509 & 0.2163 & 0.1913 & 36.49 & 0.9716 & 0.1392 & 0.1217 \\
    PDCRN\confsub{ECCV}{2020}~\cite{panikkasseril2020PDCRN} & 4.7 & 30.34 & 32.44 & 0.9559 & 0.1897 & 0.1608 & 37.83 & 0.9780 & -- & -- \\
    UDC\mbox{-}UNet\confsub{ECCV}{2022}~\cite{liu2022udcunet} & 5.7 & 338.87 & 26.25 & 0.8642 & 0.2603 & 0.2063 & 35.44 & 0.9315 & 0.1144 & 0.1462 \\
    FSI\confsub{ICCV}{2023}~\cite{liu2023fsi} & 5.3 & 155.93 & 33.14 & 0.9546 & 0.1948 & \prevsota{0.1458} & 38.21 & 0.9789 & 0.0991 & 0.1006 \\
    BNUDC\confsub{CVPR}{2022}~\cite{koh2022bnudc} & 4.6 & 496.70 & \prevsota{33.39} & \prevsota{0.9610} & \prevsota{0.1748} & 0.1511 & \prevsota{38.22} & \prevsota{0.9798} & \prevsota{0.0988} & \prevsota{0.0964} \\
    \rowcolor{RowGray}
    % Ours & 3.2 & 151.13 & \sota{33.78} & \sota{0.9626} & \sota{0.1713} & \sota{0.1425} & \sota{38.34} & \sota{0.9802} & \sota{0.0975} & \sota{0.0901} \\
    Ours & 3.2 & 151.13 & \sota{33.81} & \sota{0.9625} & \sota{0.1718} & \sota{0.1440} & \sota{38.37} & \sota{0.9802} & \sota{0.0933} & \sota{0.0897} \\
    \bottomrule[0.5pt]
    \end{tabular}%
    }
    \label{tab:maintable_Test}

    \small
    \centering
     \caption{
    \textbf{Validation results on the POLED and TOLED validation datasets}. 
     Average PSNR, SSIM, LPIPS, and DISTS are reported, with the 
    \sota{best} and \prevsota{second-best} scores colored 
    ($\uparrow$ higher is better; $\downarrow$ lower is better).
    }
    \resizebox{\textwidth}{!}{%
    \begin{tabular}{l|cc|cccc|cccc}
    \toprule[0.5pt]
    \rowcolor{HeaderBlue}
    \multirow{2}{*}{Method} & \multirow{2}{*}{\makecell[c]{\#Params\\(M)}} & \multirow{2}{*}{\makecell[c]{FLOPs\\(G)}} & \multicolumn{4}{c|}{POLED-Validation~\cite{zhou2021image}} & \multicolumn{4}{c}{TOLED-Validation~\cite{zhou2021image}} \\
    \rowcolor{HeaderBlue}
    \multirow{-2}{*}{Method} & \multirow{-2}{*}{\makecell[c]{\#Params\\(M)}} & \multirow{-2}{*}{\makecell[c]{FLOPs\\(G)}} & PSNR$\uparrow$ & SSIM$\uparrow$ & LPIPS$\downarrow$ & DISTS$\downarrow$ & PSNR$\uparrow$ & SSIM$\uparrow$ & LPIPS$\downarrow$ & DISTS$\downarrow$ \\
    \midrule[0.5pt]
    MSUNet\confsub{CVPR}{2021}~\cite{zhou2021image} & 8.9 & 124.36 & 29.96 & 0.9343 & 0.2281 & 0.1774 & 38.25 & 0.9772 & 0.1174 & 0.1155 \\
    DAGF\confsub{ECCV}{2020}~\cite{sundar2020deep} & 1.1 & 36.77 & 33.79 & 0.9580 & 0.2250 & 0.1942 & 37.87 & 0.9753 & 0.1367 & 0.1266 \\

    UDC\mbox{-}UNet\confsub{ECCV}{2022}~\cite{liu2022udcunet} & 5.7 & 338.87 & 26.79 & 0.8675 & 0.2741 & 0.2120 & 35.71 & 0.9310 & 0.1236 & 0.1524 \\
    FSI\confsub{ICCV}{2023}~\cite{liu2023fsi} & 5.3 & 155.93 & \prevsota{34.71} & 0.9576 & \prevsota{0.1869} & \sota{0.1431} & 38.96 & 0.9802 & \sota{0.1015} & 0.1104 \\
    BNUDC\confsub{CVPR}{2022}~\cite{koh2022bnudc} & 4.6 & 496.70 & 34.39 & \prevsota{0.9634} & 0.1871 & 0.1612 & \prevsota{39.09} & \prevsota{0.9814} & 0.1072 & \prevsota{0.1052} \\
    \rowcolor{RowGray}
    Ours & 3.2 & 151.13 & \sota{34.74} & \sota{0.9640} & \sota{0.1839} & \prevsota{0.1530} & \sota{39.17} & \sota{0.9816} & \prevsota{0.1034} & \sota{0.0976} \\
    \bottomrule[0.5pt]
    \end{tabular}%
    }
    \label{tab:maintable_Val}
    \end{table*}
\subsection{Loss Functions}

\paragraph{High-Frequency Uncertainty-Driven Loss.}
% UDL 설명 -> 문제점(한계) -> (한계 해결을 위한) HF-UDL 소개 -> 구현 방법 (라플라시안 사용) -> 식 정의 -> 식 설명 -> multi-stage decoder로 적용 -> HF-UDL 장점
Previous uncertainty-driven losses (UDL)~\cite{ning2021uncertainty,chen2023sparse} estimate pixel-wise uncertainty to adaptively weight the supervision, encouraging the model to focus on larger reconstruction errors. 
However, directly applying UDL is insufficient for recovering high-frequency details affected by compounded degradations such as blur, noise, and low transmittance that frequently occur in UDC images, as illustrated in Fig.~\ref{fig:loss_ablation} (b). 
To address this limitation, we extend the $\mathcal{L}_{\text{UDL}}$ in Eq.~\ref{eq:Loss3} and propose the high-frequency uncertainty-driven loss (HF-UDL) which is specifically to enhance high-frequency details. 

HF-UDL applies a Laplacian operator to capture uncertainty in high-frequency components, enabling the network to restore accurate structures under uncertainty-aware learning. 
The loss is defined as follows:
\begin{equation}
    \mathcal{L}_\text{HF-UDL}= \exp(-s)\,\big\|\Delta(\hat{I}) - \Delta(I_{gt})\big\|_1+ 2s,
    \label{eq:Loss4}
\end{equation}
where $\Delta$ is the Laplacian operator, $\hat{I}$ and $I_{gt}$ denote the restored and ground-truth images, respectively, and $s$ corresponds to the uncertainty predicted by the associated decoder block. 
Following ~\cite{chen2023sparse}, HF-UDL is applied to each decoding block to encourage uncertainty-aware feature refinement across multiple stages as shown in Fig.~\ref{fig:UDLoss}. 
The proposed loss not only enhances fine-detail but also encourages each variance estimator to learn uncertainty of high-frequency features, thereby producing more reliable uncertainty maps, as shown in Fig.~\ref{fig:uncertaintymap_main}. 

In addition, we employ the conventional PSNR loss ($\mathcal{L}_{\text{PSNR}})$ as in~\cite{huynh2008scope}, which measures the pixel-wise reconstruction fidelity between the restored and ground-truth images, and the final loss for our UDC restoration is given as follows:
\begin{equation}
    \mathcal{L}_\text{total} = \lambda_1\mathcal{L}_\text{HF-UDL} + \lambda_2\mathcal{L}_\text{PSNR},
    \label{eq:Loss7}
\end{equation}
where the weighting values of $\lambda_1$ and $\lambda_2$ are determined empirically as 100 and 0.5.
    \begin{table}[]
    \small
    \centering
    \caption{
    \textbf{Results on the SYNTH dataset}. 
    Average PSNR, SSIM, LPIPS, and DISTS are reported, with the 
    \sota{best} and \prevsota{second-best} scores colored 
    ($\uparrow$ higher is better; $\downarrow$ lower is better).
    }
    \vspace{-7px}
    \resizebox{\linewidth}{!}{%
    \begin{tabular}{l|c|cccc}
    \toprule[0.5pt]
    \rowcolor{HeaderBlue}
    \multirow{2}{*}{Method} & \multirow{2}{*}{\makecell[c]{\#Params\\(M)}} & \multicolumn{4}{c}{Synthetic Dataset} \\
    \rowcolor{HeaderBlue}
    \multirow{-2}{*}{Method} & \multirow{-2}{*}{\makecell[c]{\#Params\\(M)}} & PSNR$\uparrow$ & SSIM$\uparrow$ & LPIPS$\downarrow$ & DISTS$\downarrow$ \\ 
    \midrule[0.5pt]
    SFTMD~\cite{gu2019blind} &3.9 &42.35 &0.9863 &0.0123 & - \\
    DISCNet~\cite{DISCNet} &3.8 &43.27 &0.9877 &\sota{0.0108} &0.0182 \\
    UDC-UNet~\cite{liu2022udcunet} &5.7 &45.37 &0.9898 &0.0162 &0.0175 \\
    BNUDC~\cite{koh2022bnudc} &4.6 &45.56 &\prevsota{0.9940} &\prevsota{0.0110} &\prevsota{0.0155}\\
    FSI~\cite{liu2023fsi} &5.3 &\prevsota{45.69} &0.9930 &0.0126 &0.0206\\
    % Ours &3.2 &\sota{46.16} &\sota{0.9945} &\prevsota{0.0112} &\sota{0.0150} \\
    Ours &3.2 &\sota{46.71} &\sota{0.9942} &\prevsota{0.0110} &\sota{0.0150} \\
    \bottomrule[0.5pt]
    \end{tabular}%
    }
    \label{tab:maintable_syn}
    \end{table}

\section{Experiments}
\subsection{Experimental Setup}

\begin{figure*}[h]
    \centering
    \includegraphics[width=1.0\linewidth]{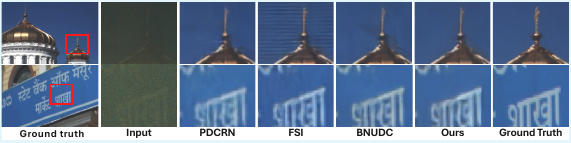}
    \caption{Qualitative comparisons on the POLED dataset.}
    \label{fig:poled_main}
    \centering
    \includegraphics[width=1.0\linewidth]{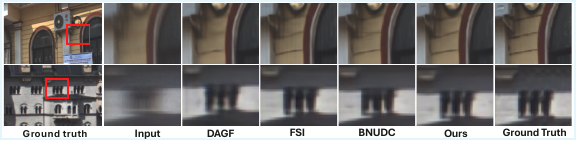}
    \caption{Qualitative comparisons on the TOLED dataset.}
    \label{fig:toled_main}
\end{figure*}

\paragraph{Implementation Details.}

In our implementation, we train our model using the Adam optimizer~($\beta_1=0.9$, $\beta_2=0.999$) and learning rate is initialized to $2\times10^{-4}$ and decayed linearly over 1,000 training epochs. 
In the training phase, Gaussian noise with a standard deviation of $\sigma = 10^{-3}$ is added to the input images to enhance robustness, whereas the ground-truth images are kept free from noise.
The entire framework is implemented in PyTorch~\cite{paszke2019pytorch} and trained on an NVIDIA RTX 3090 GPU.
The source code will be released upon publication. 

\paragraph{Datasets and Compared Methods.}
For fair comparison with existing methods, we evaluate our model on three widely used datasets: POLED~\cite{zhou2021image}, TOLED~\cite{zhou2021image}, and SYNTH~\cite{DISCNet}.
The POLED and TOLED datasets each contain 300 paired images (240/30/30 for training/validation/testing) with a resolution of 1024×2048.
Specifically, POLED images suffer from severe color shifts and low-light noise due to their low transmittance ($\sim$3\%), whereas TOLED images primarily exhibit large blurs caused by higher transmittance ($\sim$20\%). 
The SYNTH dataset consists of 2,016 training and 360 testing image pairs synthesized using PSFs computed from a commercial UDC panel, each with a resolution of 800×800. 

We compare the proposed model with existing state-of-the art UDC image restoration methods, including MSUNet~\cite{zhou2021image}, DAGF~\cite{sundar2020deep}, PDCRN~\cite{panikkasseril2020PDCRN}, UDC-UNet~\cite{liu2022udcunet}, DISCNet~\cite{DISCNet}, BNUDC~\cite{koh2022bnudc}, and FSI~\cite{liu2023fsi}.
We present results of these approaches using their official codes. 
Quantitative evaluation is conducted using PSNR~\cite{huynh2008scope} and SSIM~\cite{wang2004image} to assess pixel-level reconstruction fidelity, while LPIPS~\cite{zhang2018unreasonable} and DISTS~\cite{ding2020image} are employed to measure perceptual and texture similarity, respectively.
Note that, unlike previous UDC methods~\cite{koh2022bnudc,liu2023fsi}, UCMNet restores POLED and TOLED inputs directly without handcrafted pre-processing, demonstrating strong robustness and general applicability. 

\subsection{Performance Evaluation}
\paragraph{Quantitative Comparison.} 

Tables~\ref{tab:maintable_Test} and~\ref{tab:maintable_Val} show the performance of the proposed UCMNet on the test and validation sets of two representative UDC datasets, POLED~\cite{zhou2021image} and TOLED~\cite{zhou2021image}.
As shown in the test set results in Table~\ref{tab:maintable_Test}, our method achieves the best overall performance on both the TOLED and POLED test sets across reconstruction metrics (PSNR, SSIM), perceptual metric (LPIPS) and texture metric (DISTS). 
Similarly, the validation set results in Table~\ref{tab:maintable_Val} show that our UCMNet outperforms existing approaches on POLED across PSNR, SSIM, and LPIPS, while ranking second in DISTS.
For TOLED, ours consistently attains the highest scores in PSNR, SSIM, and DISTS, and ranks second in LPIPS. 
In addition, Table~\ref{tab:maintable_syn} UCMNet also outperforms previous method at the SYNTH~\cite{DISCNet} dataset. 
Overall, UCMNet consistently improves PSNR, SSIM, LPIPS, and DISTS, confirming that the proposed UPT effectively addresses degradations arising from under-display imaging conditions. 

\paragraph{Qualitative Comparison.}
In Fig.~\ref{fig:poled_main}, qualitative comparisons show that previous methods struggle to restore POLED-degraded images, frequently suffering from light-scattering artifacts that lead to significant detail loss in the restored results. 
Notably, PDCRN fails to resolve feature blurriness, while FSI enhances texture details, but struggles to accurately recover fine structures due to horizontal pattern artifacts.
Similarly, BNUDC exhibits noticeable errors around the boundary between the roof and the background and fails to reconstruct the text correctly. 
In contrast, the proposed method demonstrates improved restoration performance in UDC tasks characterized by regionally varying degradations through an uncertainty-driven approach, effectively removing distinctive artifacts specific to such degradations.
This effectiveness is evidenced by the clearer reconstruction of roof edges and text regions. 
For the TOLED dataset, we provide visual comparisons in Fig.~\ref{fig:toled_main}. 
TOLED-degraded images suffer from spatially varying PSFs, resulting in irregular blurs, which pose challenges for existing methods and lead to limited restoration performance. 
In contrast, ours effectively reconstructs wall patterns and clearly separated window structures, demonstrating superior detail preservation compared to other methods. 

\begin{figure}[h]
    \centering
    \includegraphics[width=0.9\linewidth]{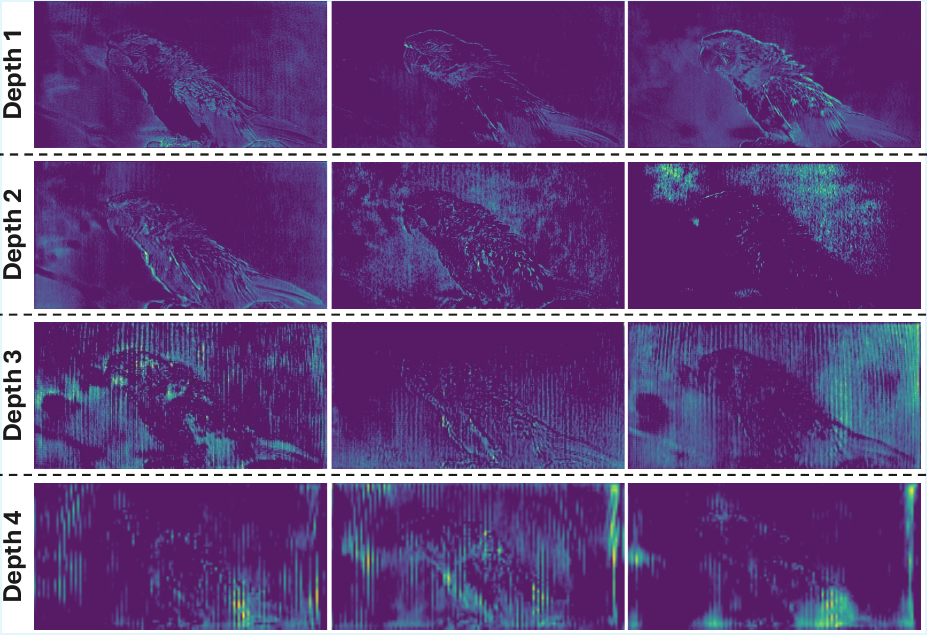}
    \vspace{-10px}
    \caption{
    Visualization of uncertainty maps across decoder stages. 
    }
    \label{fig:uncertaintymap_main}
% \end{figure}

% \begin{figure}[h]
    \centering
    \includegraphics[width=1.0\linewidth]{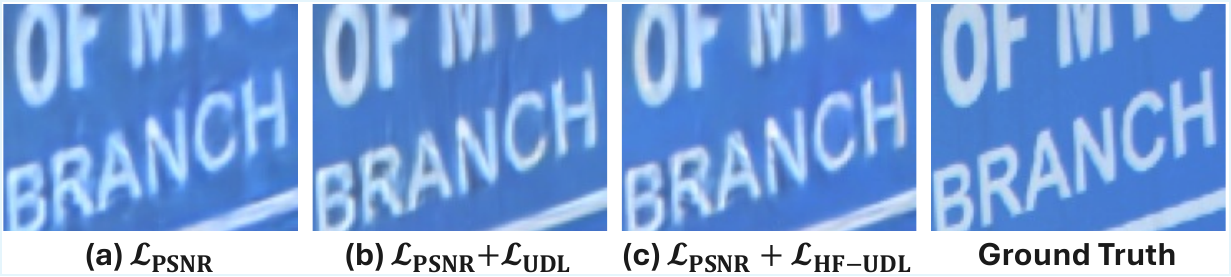}
    \caption{The visualization of the loss function ablation study, where (a), (b), and (c) correspond to the cases in Table~\ref{tab:lossablation}.}
    \label{fig:loss_ablation}
% \end{figure}

% \begin{figure}[h]
    \centering
    \includegraphics[width=1.0\linewidth]{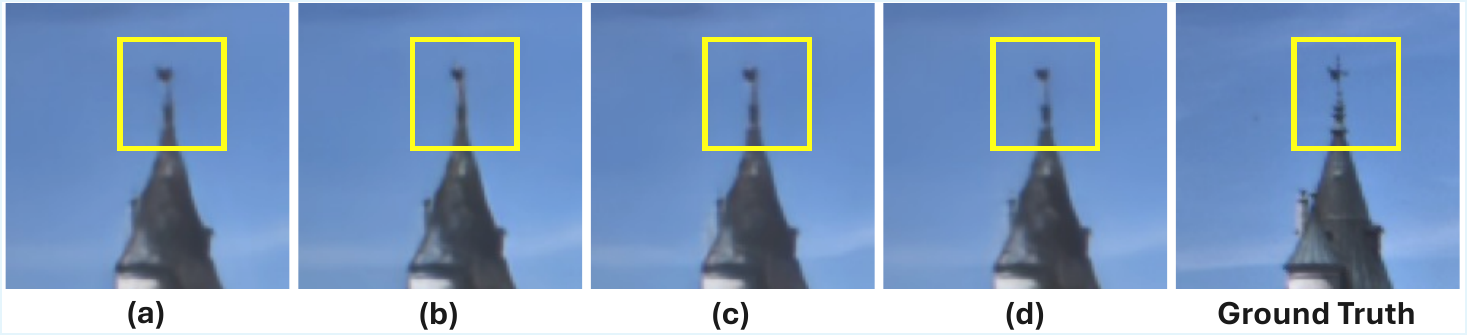}
    \caption{The visualization of the UPT ablation study, where (a), (b), (c), and (d) correspond to the cases in Table~\ref{tab:networkablation}.}
    \label{fig:UPT_ablation}
\end{figure}

\renewcommand{\arraystretch}{1.25}
\begin{table}[h]
\small
\centering
\caption{Ablation study of loss functions on POLED~\cite{zhou2021image} dataset.}
\resizebox{0.8\linewidth}{!}{%
\begin{tabular}{l|cccc}
\toprule[0.5pt]
\rowcolor{HeaderBlue}
Case & $\mathcal{L}_\text{PSNR}$ & $\mathcal{L}_\text{UDL}$ & $\mathcal{L}_\text{HF-UDL}$ & PSNR/SSIM \\ 
\midrule[0.5pt]
(a)  &\cm  &  &  &33.38/0.9600 \\
(b)  &\cm  &\cm  &  &33.59/0.9619 \\
(c)  &\cm  &  &\cm  & \textbf{33.81}/\textbf{0.9625} \\
\bottomrule[0.5pt]
\end{tabular}%
}
\label{tab:lossablation}
% \end{table}

% \renewcommand{\arraystretch}{1.25}
% \begin{table}[h]
\small
\centering
\caption{Ablations of UPT on POLED dataset. The case without $hv$-attention uses two vanilla transformer modules in the UPT.} 
\vspace{-10px}
\resizebox{\linewidth}{!}{%
\begin{tabular}{l|cccc}
\toprule[0.5pt]
\rowcolor{HeaderBlue}
\multirow{2}{*}{Case} & \multirow{2}{*}{$hv$-attention} & \multicolumn{2}{c}{Cross-attention} & \multirow{2}{*}{PSNR/SSIM} \\
\rowcolor{HeaderBlue}
\multirow{-2}{*}{Case} & \multirow{-2}{*}{$hv$-attention} & Memory Bank ($\mathbf{M}$) & Context Bank ($\mathbf{C}$) & \multirow{-2}{*}{PSNR/SSIM} \\ 
\midrule[0.5pt]
(a) &  &   & &33.59/0.9618 \\
(b) &\cm  &  & &33.64/0.9621 \\
(c) &\cm  &\cm  & &33.69/0.9617 \\
(d) &\cm  &  &\cm &\textbf{33.81}/\textbf{0.9625} \\
\bottomrule[0.5pt]
\end{tabular}%
}
\label{tab:networkablation}
% \end{table}
% \renewcommand{\arraystretch}{1.25}
% \begin{table}[h]
\small
\centering
\caption{Ablation on the use of \textbf{Uncertainty-Guided Context Generation}. 
Incorporating uncertainty as guidance for Context generation improves both PSNR and SSIM, 
demonstrating effective spatial adaptation to locally varying degradations.}
\vspace{-10px}
\resizebox{\linewidth}{!}{%
\begin{tabular}{l|ccc}
\toprule[0.5pt]
\rowcolor{HeaderBlue}
Guidance & Router & Input feature ($F_{in}$) & Uncertainty feature ($F_{U}$)\\
\midrule[0.5pt]
PSNR/SSIM & 33.48/0.9620 &33.63/0.9620 &\textbf{33.81}/\textbf{0.9625}\\
\bottomrule[0.5pt]
\end{tabular}%
}
\label{tab:promptguidance}
\end{table}

\subsection{Ablation Studies}
\paragraph{Impact of Loss Functions.} 
To investigate the impact of each loss function, we conduct an ablation study on the PSNR loss $\mathcal{L}_\text{PSNR}$, the uncertainty-driven loss $\mathcal{L}_\text{UDL}$, and the high-frequency uncertainty-driven loss $\mathcal{L}_\text{HF-UDL}$, which extends $\mathcal{L}_\text{UDL}$ for UDC restoration tasks. 
As shown in Table~\ref{tab:lossablation}, adding $\mathcal{L}_\text{UDL}$ improves the PSNR by 0.21dB, while replacing $\mathcal{L}_\text{UDL}$ with $\mathcal{L}_\text{HF-UDL}$ yields an additional 0.22dB improvement.
Although $\mathcal{L}_{\text{UDL}}$ contributes to performance improvement, Fig.~\ref{fig:loss_ablation} shows that case (b), which applies $\mathcal{L}_{\text{UDL}}$, still fails to recover a clean, artifact-free image from UDC degradations. 
In contrast, case (c) with our newly designed $\mathcal{L}_{\text{HF-UDL}}$ produces results closest to the ground truth, further enhancing image sharpness and fine details. 
We further present depth-wise uncertainty maps in Fig.~\ref{fig:uncertaintymap_main}, where early stages highlight fine structural details while later stages respond to coarse degradation patterns. 
\vspace{-10px}
\paragraph{Analysis of UPT Components.}
In Table~\ref{tab:networkablation}, we analyze the components of the UPT, focusing on the impact of incorporating horizontal and vertical ($hv$)-attention and the effectiveness of cross-attention utilizing Memory Bank $\mathbf{M}$ and Context Bank $\mathbf{C}$. 
First, compared to case (a) using vanilla transformers, introducing $hv$-attention in case (b) improves both PSNR and SSIM, demonstrating that $hv$-attention enhances the restoration capability of the model.
Next, in case (c), applying cross-attention with $\mathbf{M}$ yields a similar effect to case (b), suggesting that the uncertainty information stored in the Memory Bank provides limited contribution to restoration.
Therefore, we employ context features extracted from the Context Bank, which complements the Memory Bank by enriching restoration-relevant information.
As shown in case (d), this approach further improves PSNR and SSIM compared to cases (b) and (c). 
Visual results are shown in Fig.~\ref{fig:UPT_ablation}.
\vspace{-10px}
\paragraph{Analysis of Prior Choices for Context Feature.}
In Table~\ref{tab:promptguidance}, we analyze the impact of different priors for constructing the context feature $F_C$. 
First, using a simple three-layer router to generate the context feature directly from $F_{in}$ without any prior guidance yields the lowest PSNR of 33.48 dB, confirming the necessity of a meaningful uncertainty prior. 
Subsequently, we retrieve the context feature using the input feature $F_{in}$ instead of $F_{U}$, resulting in a 0.18 dB decrease in PSNR in comparison to our model that employs the uncertainty feature $F_{U}$.  
These results indicate that the uncertainty-prior provides greater information and reliability for contextual features.

\section{Conclusion}  
We presented UCMNet, an uncertainty-driven framework for restoring images degraded by under-display cameras. 
By introducing a high-frequency uncertainty-driven loss (HF-UDL), the model learns fine-grained uncertainty that reflects diffraction-induced high-frequency degradation. 
These uncertainty maps are further exploited by the proposed Uncertainty-Prior Transformer (UPT), which retrieves degradation-aware context through learned Memory Bank and adaptively refines features via directional cross-attention.
Through this uncertainty-guided design, UCMNet effectively handles spatially varying distortions, suppresses display-induced artifacts, and recovers high-frequency structures with improved fidelity. 
Experiments on POLED, TOLED, and SYNTH benchmarks demonstrate consistent state-of-the-art performance with significantly fewer parameters. 
Our results highlight the importance of uncertainty modeling for robust and efficient UDC image restoration. 

\paragraph{Acknowledgments} 
This work was supported by 
Institute of Information \& communications Technology Planning \& Evaluation (IITP) grant funded by the Korea government (MSIT) (No.2022-0-00156, Fundamental research on continual meta-learning for quality enhancement of casual videos and their 3D metaverse transformation), 
IITP grant funded by the Korea government (MSIT) (No.RS-2020-II201373, Artificial Intelligence Graduate School Program (Hanyang University)), and the research fund of Hanyang University (HY-2026). 

{
    \small
    \bibliographystyle{ieeenat_fullname}
    \bibliography{main}
}

\clearpage
\setcounter{page}{1}
\maketitlesupplementary

\setcounter{section}{0}
\setcounter{table}{0}
\setcounter{figure}{0}
\setcounter{equation}{0}
\renewcommand\thesection{\Alph{section}}
\renewcommand{\thetable}{S\arabic{table}}
\renewcommand{\thefigure}{S\arabic{figure}}
\renewcommand{\theequation}{S.\arabic{equation}}

This supplementary material provides additional details that were omitted from the main paper due to space constraints. 
Sec.~\ref{sec:fcm} describes the architectural details of the Frequency Convolution Module, and Sec.~\ref{sec:update} explains the training-time update mechanism of the Context and Memory Banks.
In Sec.~\ref{sec:size_ablation}, we investigate the effect of varying the size of the Context and Memory Banks and identify an appropriate configuration for UCMNet.
Sec.~\ref{sec:synth} presents the loss functions and qualitative comparisons on the SYNTH dataset, while Sec.~\ref{sec:realdataset} reports the performance of UCMNet on a real-world UDC dataset.
In Sec.~\ref{sec:lowlight}, we demonstrate the capability of UCMNet for low-light image enhancement.
Finally, Sec.~\ref{sec:uncertainty} and Sec.~\ref{sec:scoremap} provide additional visualizations of uncertainty maps and memory token usage, respectively, and Sec.~\ref{sec:limitation} presents limitation examples of UCMNet for future work. 

\section{Architectural Details of FCM}
\label{sec:fcm}
In Sec. 3.1.1, to provide reliable representations for UDC image restoration, we introduce a lightweight Frequency Convolution Module (FCM), as illustrated in Fig. 3 (b) of the main manuscript. 
Inspired by NAFNet~\cite{chen2022nafnet} and DarkIR~\cite{feijoo2025darkir}, the proposed FCM is designed to perform frequency-domain feature enhancement followed by spatial attention refinement. 
In practice, FCM is integrated into every encoder and decoder blocks to enable efficient and effective restoration.

\paragraph{Frequency-domain Feature Enhancement.}
To preserve and enhance high-frequency details, FCM transforms input feature into the Fourier domain, decomposes it into phase and amplitude, and refines the amplitude while preserving phase, which retains the structural information of the input feature. 
Then, through the inverse Fourier transformation, we obtain a frequency residual that compensates the original feature.

\paragraph{Spatial Attention Refinement.}
After frequency compensation, to address spatially varying distortions we apply a spatial attention module following the NAFNet design.
Specifically, a lightweight Single Gate (SG) selectively amplifies salient spatial activations, while a Simplified Channel Attention (SCA) module enhances feature channels to emphasize consistently informative responses. 

\section{Context Bank and Memory Bank Update}
\label{sec:update}
In Sec. 3.2.1, we explain how the uncertainty map $F_U \in \mathbb{R}^{H' \times W' \times C}$ is used as a prior to retrieve the context feature $F_C \in \mathbb{R}^{H' \times W' \times C}$ from the Memory Bank $\mathbf{M} \in \mathbb{R}^{N \times C}$ and the Context Bank $\mathbf{C} \in \mathbb{R}^{N \times C}$, where $F_C$ represents high-frequency information tailored to local degradation patterns.
In this section, we detail the update mechanism for the Memory Bank $\mathbf{M}$ and the Context Bank $\mathbf{C}$ during training. 

\paragraph{Memory Bank.}
We update the memory tokens $m_i \in \mathbb{R}^{1\times C}, \, (i \in \{1, \dots, N\})$ in the Memory Bank $\mathbf{M}$ according to their similarity to the uncertainty feature vectors $f^u_j  \in \mathbb{R}^{1\times C}, \, (j \in \{1, \dots, H'W'\})$. 
Using the cosine similarity score $s_{ij}$ in Eq. 3 of the main paper, we first identify the address of the most relevant memory token for each uncertainty feature vector:

\begin{equation}
r_j = \arg\max_{i} s_{ij}, \, (i \in \{1, \dots, N\}),
\end{equation}
where $r_j$ denotes the address of the memory token most similar to the uncertainty feature vector $f^u_j$.
The corresponding memory token $m_{r_j}$ is then updated using a momentum-based rule following ~\cite{fang2025parameterized,huang2021memory} so that it progressively adapts to the uncertainty feature:
\begin{equation}
m_{r_j} \leftarrow \eta \cdot m_{r_j} + (1-\eta) \cdot f^u_j,
\end{equation}
where $\eta$ is the momentum coefficient. 

\paragraph{Context Bank.}
The tokens $c_i$ in the Context Bank $\mathbf{C}$ serve as the components of the context feature $F_C$, as described in Sec. 3.2.1.
They are updated jointly with the other network parameters during training, allowing them to gradually encode high-frequency information that supports accurate restoration of UDC images.

\section{Ablation on Context-Memory Bank Size}
\label{sec:size_ablation}
% We analyze the effect of token numbers of both Memory and Context Banks. 
In Table~\ref{tab:promptsize}, we investigate the impact of the number of tokens ($N$) in the Memory and Context Banks.
When the token number $N$ is 256, the model provides the best trade-off between representational capacity and restoration performance.

\renewcommand{\arraystretch}{0.9}
\begin{table}[h]
\centering
\caption{Ablation on the size of the Context–Memory Bank (\(N\)). 
A compact configuration (\(N{=}256\)) achieves the best balance between accuracy and efficiency.}
\resizebox{0.7\linewidth}{!}{%
\begin{tabular}{c|ccc}
\toprule[0.5pt]
\rowcolor{HeaderBlue} 
CMB size ($N$) & 128 & \textbf{256} & 512 \\
\midrule[0.5pt]
\rowcolor{white}
PSNR$\uparrow$ & 33.71 & \textbf{33.78} & 33.60 \\
SSIM$\uparrow$ & 0.9624 & \textbf{0.9626} & 0.9619 \\
\bottomrule[0.5pt]
\end{tabular}%
}
\label{tab:promptsize}
\end{table}

\begin{figure*}[]
    \centering
    \vspace{-10px}
    \includegraphics[width=1.0\textwidth]{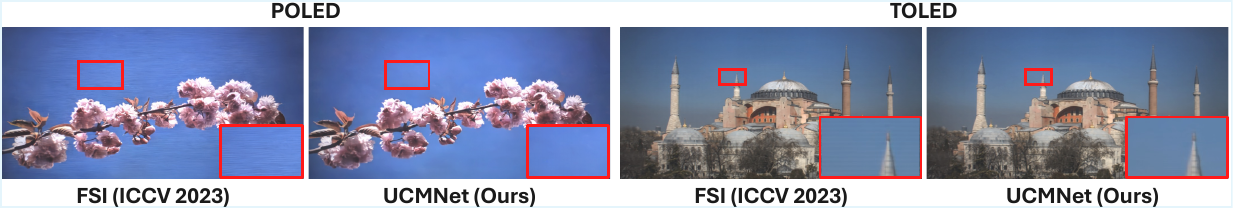}
    \caption{Please zoom in for the original-size comparison.}
    \label{fig:diffmap_full}
% \end{figure*}
% \begin{figure*}[h]
    \centering
    \includegraphics[width=1.0\linewidth]{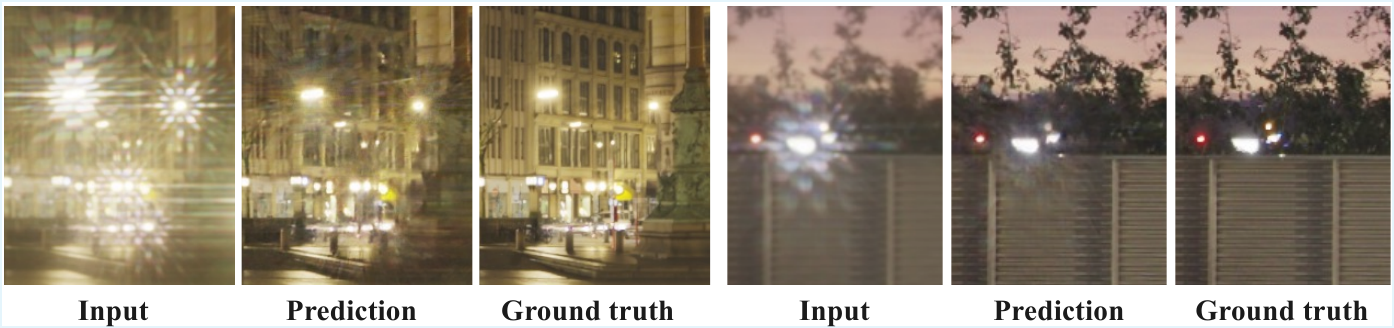}
    \caption{
    Limitation examples of UCMNet on SYNTH dataset.
    }
    % \vspace{-10px}
    \label{fig:limitation_syn}
    % \vspace{-10px}
\end{figure*}

\section{SYNTH Dataset}
\begin{figure*}[h]
    \centering
    \includegraphics[width=1.0\linewidth]{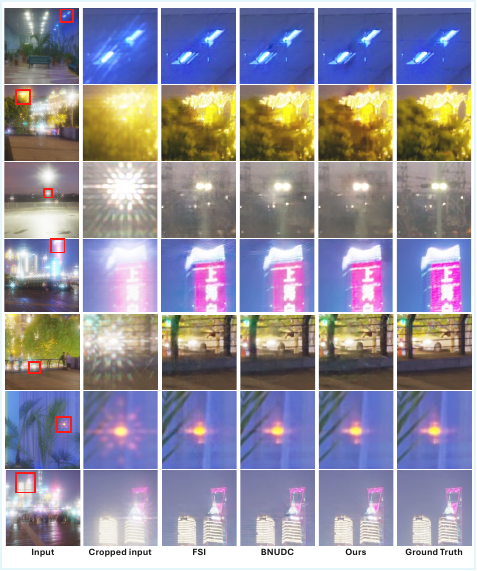}
    \caption{
    Visual results on the SYNTH dataset. 
    }
    \label{fig:synth}
\end{figure*}

\paragraph{Loss Function.}
\label{sec:synth}
Similar to BNUDC~\cite{koh2022bnudc} and FSI~\cite{liu2023fsi}, we employ the L1 loss on the SYNTH dataset rather than the PSNR loss, as follows:
\begin{equation}
    \mathcal{L}_\text{total} = \lambda_1\mathcal{L}_\text{HF-UDL} + \lambda_2\mathcal{L}_\text{L1},
    \label{eq:Loss7}
\end{equation}
where $\mathcal{L}_\text{L1}$ denotes the pixel-wise L1 loss between the prediction $\hat{I}$ and $I_{gt}$.
The weighting values of $\lambda_1$ and $\lambda_2$ are empirically determined as 100 and 0.5.
% \vspace{-20px}
\paragraph{Visualization.}
% Since in the manuscript, only TOLED and POLED visual comparisons are provieded like Fig.~\ref{fig:diffmap_full}, for qualitative comparison on the SYNTH dataset, we present visual results in Fig.~\ref{fig:synth}. 
As the manuscript includes visual comparisons only for the TOLED and POLED datasets like Fig.\ref{fig:diffmap_full}, we provide qualitative results on the SYNTH dataset in Fig.\ref{fig:synth}. 
Unlike FSI~\cite{liu2023fsi} and BNUDC~\cite{koh2022bnudc}, our method achieves noticeably better detail restoration and effectively suppresses light-diffraction artifacts characteristic of UDC imaging conditions.

\section{Real-world Dataset}
\label{sec:realdataset}
\begin{table}[t]
\centering
\resizebox{1.0\columnwidth}{!}{%
\begin{tabular}{l|cc|cc}
\hline
\rowcolor{HeaderBlue}
Dataset     & PPM-UNet  & AlignFormer+PPM-UNet  & UCMNet  & AlignFormer+UCMNet \\ \hline
AlignFormer & 19.03 / 0.7808  & 22.95 / 0.8581  & \textbf{20.41 / 0.8168}  & \textbf{23.11 / 0.8612}
  \\ \hline
\end{tabular}%
}
\vspace{-10px}
\caption{\small PSNR/SSIM results on real UDC dataset from AlignFormer~\cite{alignformer}.}
\label{tab:AlignFormer}
\end{table}
\begin{table}[t]
\centering
% \tiny
\resizebox{1.0\columnwidth}{!}{%
\begin{tabular}{l|ccccc}
\hline
\rowcolor{HeaderBlue}
Dataset     & Restormer  & GSAD & Retinexformer  & DarkIR  & UCMNet (Ours) \\ \hline
LOLv2-Real  & 19.94 / 0.827  & 20.15 / 0.846  & 22.79 / 0.840  & \textbf{23.87 / 0.880}  & \underline{23.10 / 0.862} \\
LOLv2-Syn   & 21.41 / 0.830  & 24.47 / 0.929  & \textbf{25.67} / 0.930  & \underline{25.54 / 0.934}  & \textbf{25.67 / 0.936} \\ \hline
\end{tabular}%
}
\vspace{-10px}
\caption{UCMNet on low-light image enhancement task.}
\label{tab:lowlight}
\end{table}

Although the TOLED, POLED, and SYNTH datasets simulate real-world UDC degradations, they remain synthetic. 
Therefore, in Table~\ref{tab:AlignFormer}, we further evaluate our method on the real-world dataset released by AlignFormer~\cite{alignformer}, where UCMNet also demonstrates robust performance.

\section{UCMNet on Low-light Image Enhancement}
\label{sec:lowlight} 
In this section, we explore the potential of the proposed UCMNet for another image restoration task, namely low-light image enhancement, which also involves recovering images degraded by suboptimal lighting conditions. 
As in Table~\ref{tab:lowlight}, we apply UCMNet to LOLv2-Real~\cite{lolv2} and LOLv2-Syn~\cite{lolv2} to validate its broader applicability, where degradations are spatially varying. 
Even without being fully optimized for this specific task, UCMNet achieves competitive performance compared to existing methods, indicating its potential for handling tasks with spatially varying degradations under challenging lighting conditions. 

\section{Uncertainty Map Visualization}
\label{sec:uncertainty}
\begin{figure*}[h]
    \centering
    \includegraphics[width=0.95\linewidth]{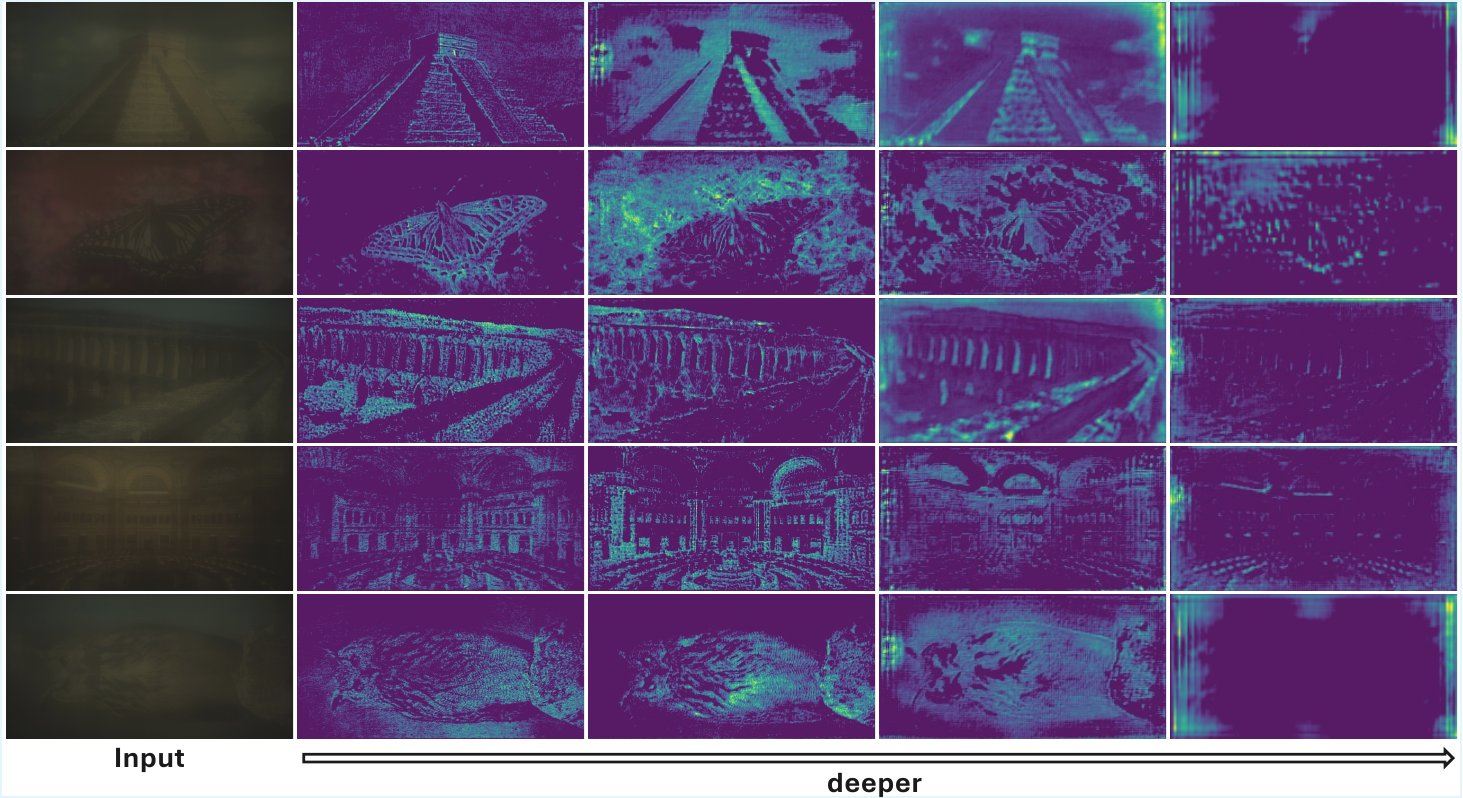}
    \caption{
    % Depth-wise visualization of estimated uncertainty maps on the POLED~\cite{zhou2021image} dataset.
    Estimated uncertainty map visual results on the POLED~\cite{zhou2021image} dataset. 
    }
    \label{fig:uncertaintymap}
\end{figure*}

As an extension of Fig. 8 in the main paper, additional uncertainty maps are provided in Fig.~\ref{fig:uncertaintymap}.
We observe that shallower stages emphasize fine structural details, whereas deeper stages primarily capture coarse degradation patterns.
This indicates that the network performs depth-wise restoration guided by the uncertainty prior. 

\section{Memory Token Usage Visualization}
\label{sec:scoremap}
\begin{figure*}[h]
    \centering
    \includegraphics[width=0.95\linewidth]{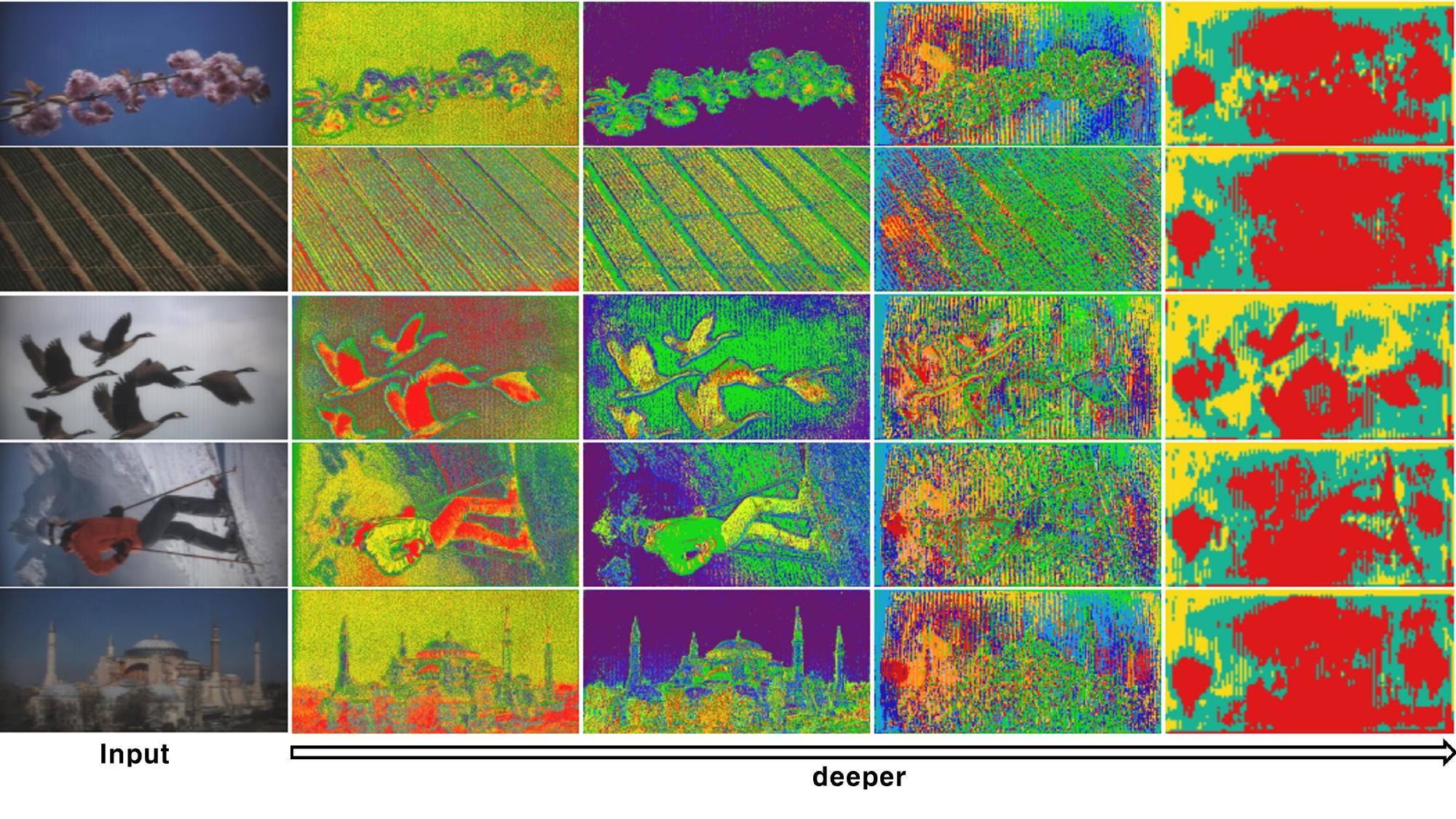}
    \caption{
    Visualized memory token usage as pseudo–segmentation maps on the TOLED~\cite{zhou2021image} dataset, where each color represents a distinct memory token associated with specific uncertainty feature vectors, allowing inspection of their spatial distribution across the image.
    }
    \label{fig:scoremap}
\end{figure*}

In Fig.~\ref{fig:scoremap}, we visualize the usage of memory tokens by representing them as pseudo–segmentation maps, where each color corresponds to a different memory token associated with  particular uncertainty feature vectors. 
This allows us to examine the spatial distribution of uncertainty features across the image.
The results are shown depth-wise across decoder stages. 
In deeper stages, the maps converge to similar patterns, indicating that uncertainty becomes focused on display-induced degradations. 
In contrast, shallower stages highlight high-frequency structural cues such as edges and contours.
These observations demonstrate that UCMNet effectively leverages diverse memory tokens to model region-specific uncertainty, enabling spatially adaptive restoration.

\section{Limitations}
\label{sec:limitation}
Although the proposed network demonstrates promising restoration performance on UDC images, UCMNet still struggles on the SYNTH dataset to recover fine details in regions with severe diffraction-induced information loss, as in Fig.~\ref{fig:limitation_syn}.
In future work, we plan to address this limitation by incorporating stronger model priors. 
% \clearpage

\end{document}